\documentclass[letterpaper, 10 pt, conference]{ieeeconf}  
\usepackage{cite}
\usepackage{algorithm}
\usepackage{algorithmic}
\usepackage{amsmath}
\usepackage{amssymb}
\usepackage{mathtools}
\usepackage{atbegshi}
\AtBeginShipoutNext{\AtBeginShipoutUpperLeft{
  \put(\dimexpr\paperwidth-2cm\relax,-1.5cm){\makebox[0pt][r]{Accepted manuscript International Conference on Rehabilitation Robotics (ICORR) 2023}}}}

\IEEEoverridecommandlockouts                              

\title{\LARGE \bf
Towards AI-controlled movement restoration: Learning FES-cycling stimulation with reinforcement learning}  

\author{Nat Wannawas$^1$,
        A. Aldo Faisal$^{1,2}$
\thanks{$^{1}$Brain \& Behaviour Lab, Imperial College London, United Kingdom, SW7 2AZ. (nat.wannawas18@imperial.ac.uk). $^2$Chair of Digital Health, Universität Bayreuth, Germany 95445. (aldo.faisal@imperial.ac.uk).}
}

\begin{document}

\maketitle
\thispagestyle{empty}
\pagestyle{empty}

\begin{abstract}
Functional electrical stimulation (FES) has been increasingly integrated with other rehabilitation devices, including rehabilitation robots. FES cycling is one of the common FES applications in rehabilitation, which is performed by stimulating leg muscles in a certain pattern. The appropriate pattern varies across individuals and requires manual tuning which can be time-consuming and challenging for the individual user. Here, we present an AI-based method for finding the patterns, which requires no extra hardware or sensors. Our method starts with finding model-based patterns using reinforcement learning (RL) and customised cycling models. Next, our method fine-tunes the pattern using real cycling data and offline RL. We test our method both in simulation and experimentally on a stationary tricycle. Our method can robustly deliver model-based patterns for different cycling configurations. In the experimental evaluation, the model-based pattern can induce higher cycling speed than an EMG-based pattern. And by using just 100 seconds of cycling data, our method can deliver a fine-tuned pattern with better cycling performance. Beyond FES cycling, this work is a case study, displaying the feasibility and potential of human-in-the-loop AI in real-world rehabilitation.
\end{abstract}
Index Terms--Electrical Stimulation, FES, Reinforcement Learning, FES cycling, stimulation pattern

\section{INTRODUCTION}
Functional Electrical Stimulation (FES) induces muscle contraction through low-energy electrical signals, allowing individuals with paralyses to perform physical exercises. FES cycling is one of the most widely performed exercises that can help prevent adverse health effects such as muscle atrophy and improve cardiovascular fitness \cite{Ferrante2008, Bo2017}. FES cycling is also integrated with robotic exoskeleton to provide gait rehabilitation training \cite{Mazzoleni2017}.

FES cycling is achieved by repetitively stimulating leg muscles in a certain pattern. This stimulation pattern plays a major role in delivering smooth and fast cycling. Different individuals require different stimulation pattern. Finding an appropriate pattern is usually carried out during the system setup. In many practical situations, finding the pattern is still a manual process based on trial-and-error and clinicians' experiences \cite{Hunt2004,Kim2008,Mcdaniel2017}. This manual process can be time-consuming and challenging, especially for a lone user who performs the cycling at home. Several methods for finding the stimulation patterns have been explored. One of them is to mimic the natural order of muscle activation which can be observed through electromyogram (EMG) of healthy subjects during voluntary cycling \cite{Hunt2004,Petrofsky2003,Wakeling2009,Lopes2014,Metani2016}. This EMG-based method is not easy to implement as it requires an experimental setup and a healthy subject with similar body size to the paralysed individual. In addition, the obtained pattern may not be optimal for the target individual \cite{Schmoll2022}. Another method is to use a biomechanical model to compute the angles where certain muscles produce positive torque. Several model-based methods based on different cycling models and cost functions were developed. Chen et al \cite{Chen1997} used a five-bar linkage cycling model and determined the pattern from the flexion and extension of hip and knee joints. Idso \cite{Idso2004} used similar linkage model but determined the pattern based on a metabolic cost. Li et al. \cite{Li2009} added muscle routing details into the linkage model and used multilayer perceptrons to approximate the angles where the muscles induce positive torque on the crank. These model-based studies, however, rely heavily on the models' accuracy, and the experimental evaluations are not thoroughly reported.

Several methods have reported successes in real-world experiments. Wiesener et Schauer \cite{Wiesener2017} proposed a method that transforms thigh and knee angles into a fixed range on which the seat-position-independent pattern can be determined. Several other methods utilise torque feedback measured through force sensors attached to the pedals. Ambrosini et al. \cite{Ambrosini2014} used both EMG and torque feedback and determined the pattern based on the regions where EMG and positive torque overlap. Maneski et al. \cite{Maneski2017} and Schmoll et al. \cite{Schmoll2022} proposed similar methods in which the pedal force was recorded from the passive (motorised or human-assisted) and active (stimulated) sessions. The angles where the stimulated muscles produce positive torque are then obtained by removing the passive force from the total force. These methods are effective but require experimental setups and cycling equipment with force sensors, which can be challenging for at-home practices.

In this work, we present a pattern-finding method that requires neither a force sensor nor motorised cycling equipment. Our method utilises reinforcement learning (RL), a machine learning algorithm that learns to do tasks by interacting with environments. RL's applications in FES controls have been explored in \cite{Wannawas2021,Wannawas2022,Fischer2021,Abreu2022,Wannawas2023a}. A closely-related study is our previous work \cite{Wannawas2021} which focuses on controlling cycling speed. The RL setup in that work, however, does not deliver the patterns that behave well on conventional FES cycling systems. Additionally, it requires a large amount of data that is difficult to gather from human-in-the-loop systems.

This work improves the setup in our previous work \cite{Wannawas2021} so as to learn well-behaved patterns and provides a strategy to learn further in the real world with a handful amount of data. 
Our method is compatible with conventional FES cycling systems and can be used by the cyclists themselves. We demonstrate our method and compare the patterns governed by our and EMG-based methods as well as their cycling performances in a real-world setting.

\section{Method}
\begin{figure*}[htbp!]
    \begin{center}
    \includegraphics[width=1.90\columnwidth]{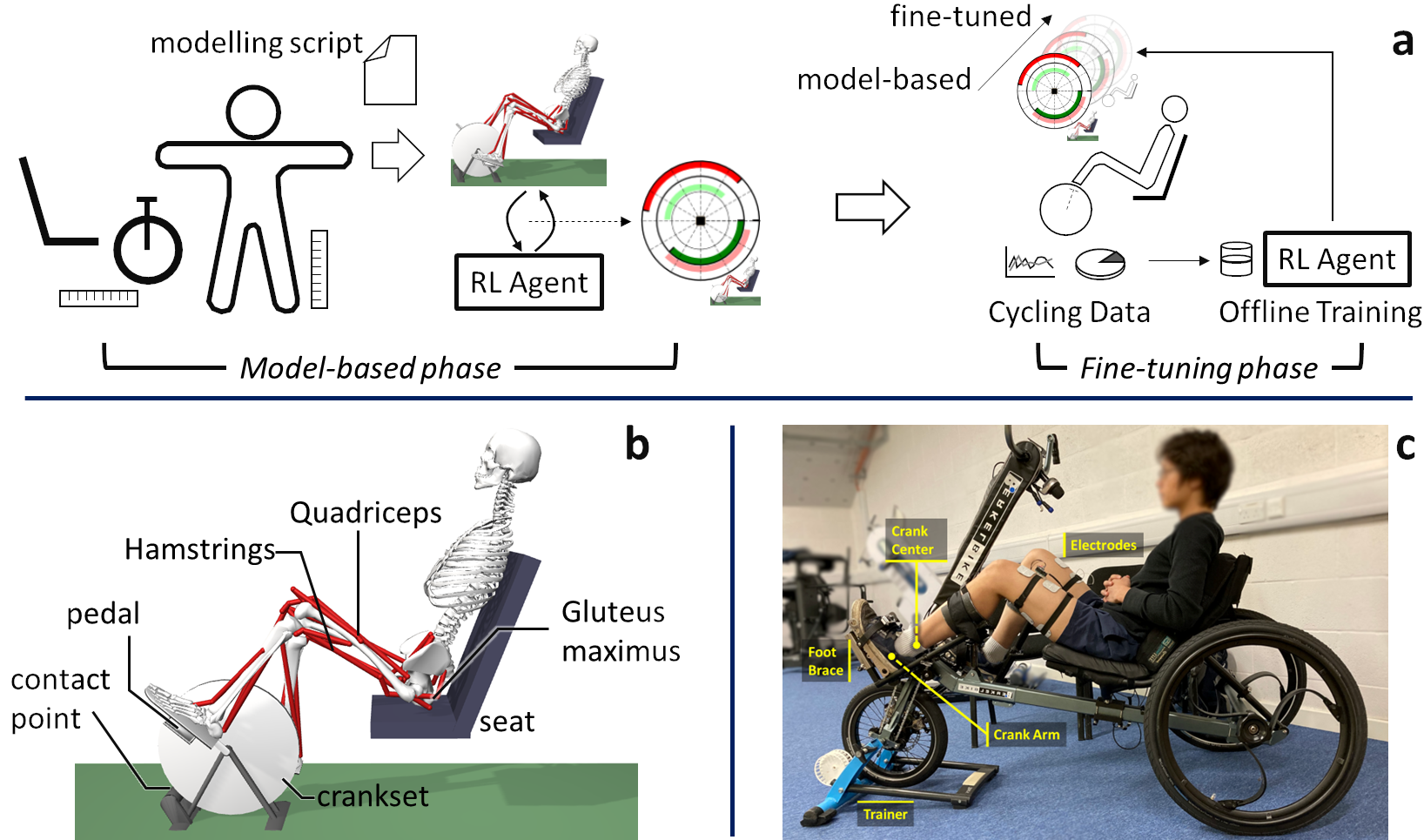}
    \caption{(a) The diagram showing the pipeline of 2-phase our method. The process starts in the model-based phase (left) that involves building a customised model and RL training to obtain a model-based pattern. The fine-tuning phase (right) involves collecting cycling data and offline RL training, causing the pattern to gradually evolve into a fine-tuned one. (b) Our OpenSim cycling model in a generic configuration. (c) The experimental setup using BerkelBike placed on a cycling trainer. The stimulation is applied to \emph{quadriceps} and \emph{hamstrings} via self-adhesive surface electrodes.
    \label{fig:Framework}}
    \vspace{-7mm}
    \end{center}
\end{figure*}

The objective of our method is to find a stimulation pattern which is crank angle intervals in which the stimulated muscles are in \emph{ON} states. Our method has two phases: model-based and fine-tuning phases (Fig.\ref{fig:Framework}a). The process starts in the model-based phase with the building of a customised musculoskeletal model corresponding to the real cyclist and cycling setup. After that, reinforcement learning (RL) is applied to find a model-based pattern. Next, in the fine-tuning phase, the process alternates between collecting cycling data and updating the pattern through offline RL. 

\subsubsection{Musculoskeletal Model}
The cycling model is built using open-source biomechanical simulation software called OpenSim \cite{Seth2011}. This software provides established components such as joints and muscles to assist the building processes. In addition, Opensim has a Python application programming interface (API) that facilitates integration with machine learning libraries. The API also allows the model to be customised through a single command, thereby obviating the need for users' skill in using OpenSim.

Fig.\ref{fig:Framework}b shows our OpenSim cycling model in a generic configuration. The model has a cycling crankset placed on a cycling trainer that exerts rolling resistance through friction at the contact point. The pedals of the crankset are attached to the feet of a lower-limb musculoskeletal model, an OpenSim built-in model, sitting on a seat. The lower-limb model has 18 Hill-type muscles; 6 of them which are \emph{quadriceps}, \emph{hamstrings}, and \emph{gluteus maximus} on both legs are stimulated. The muscles' activation delay is changed from the default value of 50 ms to 100 ms to capture the delay of FES-induced activation which is longer than natural activation \cite{Kralj1973}. We also re-route the \emph{quadriceps} muscles and modify the knee joint of the original built-in model so that the legs' lengths can be easily adjusted to match a real person. The lower-limb model has 4 movable joints: the hip and knee joints of both legs. The ankle joints are fixed at $90^\circ$. The lumbar joint, which is equal to the seat angle, is adjustable. This cycling model can be customised to match the real cyclist and cycling setup. The customisation parameters are the vertical and horizontal distances between the crank centre and the hip, the crank arm's and legs' lengths, and the seat angle.

\subsubsection{Reinforcment Learning (RL)}
Here, we use RL, a machine learning algorithm with a learning agent that learns to control an environment by interacting with it, to learn the stimulation pattern. The interactions occur in a discrete-time fashion, described as follows. At the beginning of each timestep, the agent observes an environment's state $\textbf{s}$ and selects an action $\textbf{a}$ based on its policy $\pi$. The action is applied to the environment, causing its to be in a new state $\textbf{s}'$. The agent then receives an immediate reward $r$ and observes the new state. This interaction experience is collected as a tuple of $\textbf{s}, \textbf{a}, r, \textbf{s}'$ which is stored in a replay buffer $\mathcal{D}$. This experience tuple is used to learn an optimal policy $\pi^*$ that maximises a return $R$--the sum of discounted rewards.

The learning mechanisms are different across different RL algorithms. Here, we choose an algorithm called Soft Actor-Critic (SAC) \cite{Haarnoja2019_1}, one of the state-of-the-art RL algorithms with successes in real-world control tasks. SAC has two components: an actor and a critic. In simple terms, the critic learns to estimate the expected return of a state-action pair, known as Q value ($Q(\boldsymbol{s},\boldsymbol{a})$). The Q value is used to adjust the actor's policy $\pi$ by increasing the probability of choosing an action with a high Q value. Both actor and critic are usually parameterised by neural networks with parameters $\boldsymbol{\phi}$ and $\boldsymbol{\psi}$, respectively. In SAC, the parameters $\boldsymbol{\psi}$ and $\boldsymbol{\phi}$ are optimise with gradient descent to minimise cost functions
\begin{equation}
    \label{eq:q}
    \begin{split}
        J_Q =\;&\mathbb{E}_{s,a,r,s'\sim\mathcal{D}}[Q_\psi(s,a) - (r(s,a)\\&+\gamma Q_\psi(s,\pi(s')))-log(\pi_\phi(a'|s'))]^2,
    \end{split}
\end{equation}
where $\gamma\in[0,1)$ is a discount factor, and
\begin{equation}
    \label{eq:p}
    \begin{split}
        J_\pi =\;&\mathbb{E}_{s\sim\mathcal{D}}[\mathbb{E}_{a\sim\pi_\phi}[log(\pi_\phi(s,a))-Q_\psi(s,a)]].
    \end{split}
\end{equation}

\subsubsection{RL problem formulation}
The process of learning model-based patterns involves the interaction with the OpenSim cycling model whose state $\boldsymbol{s}$ comprises the crank angle $\theta_c$ in $rad$ and the cadence $\dot{\theta}_c$ in $rad/s$. The cycling movement is controlled through the stimulation on leg muscles: \emph{quadriceps}, \emph{hamstrings}, and \emph{gluteus maximus}. The control vector, which contains the normalised stimulation intensities, is $\boldsymbol{u}\in\mathbb{R}^6$, $u_i \in [0,1]$. Note that in a two-muscle cycling case, \emph{gluteus maximus} are excluded, and $\boldsymbol{u}$ has 4 elements.

The setup on the RL agent side is as follows. We aim to govern the patterns that symmetrically stimulate right and left legs, the kind of patterns commonly used in practices and studies (e.g. \cite{Bo2017}). One strategy to achieve that is to govern the pattern based on only one leg and make the pattern for the other leg by rotating the governed one by $180^\circ$. This strategy tremendously reduces the search space and consequently decreases the required amount of data and computing time. Using this strategy, we govern the pattern with respect to the right leg. Hence, the action vector at time step $t$ becomes $\boldsymbol{a}_t\in\mathbb{R}^3$ ($a_i \in [0,1]$) for the three-muscle case and $\boldsymbol{a}_t\in\mathbb{R}^2$ for the two-muscle case. The observed state vector $\boldsymbol{s}_t$ is $[sin(\theta_{c,t}), cos(\theta_{c,t}), \dot{\theta}_{c,t}, \boldsymbol{a}_{t-1}]$, where $\theta_c$ is the right crank angle, and $\dot{\theta}_{c,t}$ is the crank's angular velocity. We represent the crank angle in the form of its sine and cosine because we want the states at $\lim_{\theta_c\to2\pi+}\theta_c$ and $\lim_{\theta_c\to2\pi-}\theta_c$ to be close in the state space and smoothly continue after each full revolution. The reward function is
\begin{equation}
r_t=f_{\mbox{reward}}(\dot{\theta}_{c,t+1}, \boldsymbol{a}_t) = \dot{\theta}_{c,t+1} - \beta\sum_{i=1}^n a_{i,t}^2,
\label{eq:reward}
\end{equation}
where $\dot{\theta}_{c,t+1}$ is the cadence at the next time step after $a_t$ was applied; $a_{i,t}$ is the stimulation on muscle $i$ at time $t$; $\beta$ is an action penalty weight which is set to $1.0$; and $n$ is the number of stimulated muscles. The intuition of the reward function is that we want to find the stimulation pattern that produces forward torque on the crank. Given a constant rolling resistance, high cadence ($\dot{\theta}_{c,t+1}$) should be observed if the forward torque is effectively induced in each revolution. The action penalty term encourages the efficiency of stimulation to minimise muscle fatigue.

The interaction between the RL agent and the cycling model is as follows. The full control vector for each timestep $\boldsymbol{u}_t$ is obtained by concatenating the action vector for right ($\boldsymbol{a}_{\mbox{right},t}$) and left ($\boldsymbol{a}_{\mbox{left},t}$) legs, which are obtained by
\begin{equation}
    \begin{split}
        \boldsymbol{a}_{\mbox{right},t} &= \pi_{\phi}(\boldsymbol{s}_t),\\
        \boldsymbol{a}_{\mbox{left},t} &= \pi_{\phi}(\boldsymbol{s}_{\mbox{left},t}),
    \end{split}
\end{equation}
where
\begin{equation}
    \begin{split}
        \boldsymbol{s}_t &= [sin(\theta_{c,t}),cos(\theta_{c,t}),\dot{\theta}_{c,t},\boldsymbol{a}_{\mbox{right},{t-1}}],\\
        \boldsymbol{s}_{\mbox{left},t} &= [-sin(\theta_{c,t}),-cos(\theta_{c,t}),\dot{\theta}_{c,t},\boldsymbol{a}_{\mbox{left},t-1}].
    \end{split}
\end{equation}
Following this, we can obtain two immediate rewards, $r_{\mbox{right},t}=f_{\mbox{reward}}(\dot{\theta}_{c,t+1},\boldsymbol{a}_{\mbox{right},t})$ and $r_{\mbox{left},t}=f_{\mbox{reward}}(\dot{\theta}_{c,t+1},\boldsymbol{a}_{\mbox{left},t})$, and two experience tuples, $(\boldsymbol{s}_t,\boldsymbol{a}_t,r_{\mbox{right},t},\boldsymbol{s}_{t+1})$ and $(\boldsymbol{s}_{\mbox{left},t},\boldsymbol{a}_{\mbox{left},t},r_{\mbox{left},t},\boldsymbol{s}_{\mbox{left},t+1})$, from an interaction in one timestep.

The RL training is episodic. Each episode starts at a random crank angle with zero initial cadence. Each episode has 100 timesteps with a size of 50 ms. The agent's parameters $\boldsymbol{\phi}$ and $\boldsymbol{\psi}$ are updated at the end of each episode. The performance test episode, which is an episode without the random actions for exploration, is carried out every 5 episodes. The training stops when the performance reaches a plateau. The stimulation pattern (\emph{On}/\emph{Off} angles) is then obtained by applying a threshold of $0.5$ to the learned continuous policy. Based on our empirical experiments, the results are not very sensitive to the threshold value because the policies change rapidly from low stimulation (near $0$) to high stimulation (near $1$) at \emph{On} angles.

\subsubsection{Fine-tuning through offline RL}
The RL setup described earlier has an RL agent interacting directly with the environment (the OpenSim cycling model). In real-world settings, however, the direct interaction may have some issues. Firstly, direct interaction requires low-latency interfaces between stimulators and computers. This may not be the case for some stimulators with built-in control units that are designed to run specific lightweight programs. Secondly, the direct interaction may have a safety issue because the RL agent can apply any form of stimulation in any situation to explore their outcomes. This can result in, for example, a sudden stop of movement, causing injury.

To avoid these issues, we collect the data from cycling sessions with conventional patterns that have a single \emph{ON} interval for each muscle. The collected data are then converted into experience tuples in the same way as described earlier. This learning setting, in RL context, is called offline learning, a learning setting where an RL agent learns from a fixed set of experiences that are not collected by the agent's policy. Offline learning poses challenges in learning an optimal policy as, for example, the agent may think that actions that have never been executed are good.

To successfully perform offline learning, we adopt one of the state-of-the-art offline RL algorithms called conservative Q learning (CQL) \cite{Kumar2020}. CQL provides the modification of RL algorithms to successfully perform offline learning at a minimal extra computational cost. CQL, in brief, learns conservative Q values by allowing the values of only state-action pairs that were applied during the data collection to be high. This is done by adding a regularisation term to the Q objective function (Eq.\ref{eq:q}) as
\begin{equation}
    \label{eq:cql}
    \begin{split}
        J_{CQL} =\;&\mathbb{E}_{s\sim\mathcal{D}}[log\sum_a exp(Q_\psi(s,a))\\
        &-\mathbb{E}_{a\sim\pi_\beta(a|s)}[Q_\psi(s,a)]]+\frac{1}{2}J_Q,
    \end{split}
\end{equation}
where $\pi_\beta$ is the policy that collected the data, which is the model-based pattern in our case.

\subsubsection{RL Architecture}
In this work, the actor and critic are parameterised by neural networks with two hidden layers. The hidden layers have 64 units with ReLU activation function. Note that the number of hidden units were determined empirically, starting the empirical search at 250 units \cite{Wannawas2021} and reducing the number until the agent fails to learn the pattern. This is to avoid over-fitting and improve the learning speed. Sigmoid activation function is applied at the policy network's output layer to squash the output between $[0,1]$, making it compatible OpenSim's muscle activation.

\subsection{Experiment}
We carry out two sets of experiments: model-based and real-world. The former is for evaluating the robustness. Note that the learning of the RL agent is a stochastic process where the success may depend on the initialisation or happen by chance. The latter shows how a model-based pattern evolves into a fine-tuned pattern with improved performance.

\subsubsection{Model-based experiments}
In these experiments, we apply our method to 10 different cycling configurations (different seat positions and legs' length) in both two-muscle and three-muscle cycling settings, totalling 20 test cases. 

\subsubsection{Real-world experiment}
We demonstrate the full pipeline of our method in a real-world setting. The experiment was carried out on a healthy subject (male, 30). This allows us to collect an exact corresponding EMG-based pattern for performance comparisons. The experiment had the subject performing two-muscle (\emph{quadriceps} and \emph{hamstrings}) cycling on a tricycle (BerkelBike, BerkelBike BV, Sint-Michielsgestel, The Netherlands), placed on a cycling trainer (Fig.\ref{fig:Framework}c). FES pulses were generated by RehaStim1 (HASOMED GmbH, Magdeburg, Germany) and delivered to the muscles via self-adhesive electrodes. At the beginning of the experiment, the right leg's length of the subject and the distance between the seat and the crank centre were measured; and the OpenSim cycling model was tailored accordingly. After that a model-based pattern was governed through our RL-based method.

Next, we applied the model-based pattern on the real cycling setup and collected cycling data. We performed ten 10-second cycling sessions with slightly different patterns obtained through simple adjustments such as shrinking, extending, and rotating the \emph{ON} intervals. This yielded 100 seconds of cycling data, which was equivalent to 4,000 experience tuples. The RL agent was then trained on these tuples in the offline mode, and a fine-tuned pattern is obtained. The performances of the model-based, fine-tuned, and EMG-based patterns were tested on 30-second cycling sessions, with 30-minute breaks between each session to minimise the effects of muscle fatigue.

The EMG-based pattern was collected by recording the EMG from the right \emph{quadriceps} and \emph{hamstrings} of the subject while performing voluntary cycling. The EMG data were collected through Olimex EKG-EMG shield (Olimex Ltd, Bulgaria) with the sampling frequency of 500 Hz and 5th-order Butterworth filter (10-250Hz cutoff) \cite{Xiloyannis2017}. The raw data were processed into root mean square (RMS) values, which show the muscle activation. The EMG pattern was then obtained by averaging the crank angle intervals where the muscles were voluntarily active.

\section{Results}
\subsection{Model-based experiments}
\begin{figure}[!hb]
    \begin{center}
    \includegraphics[width=0.99\columnwidth]{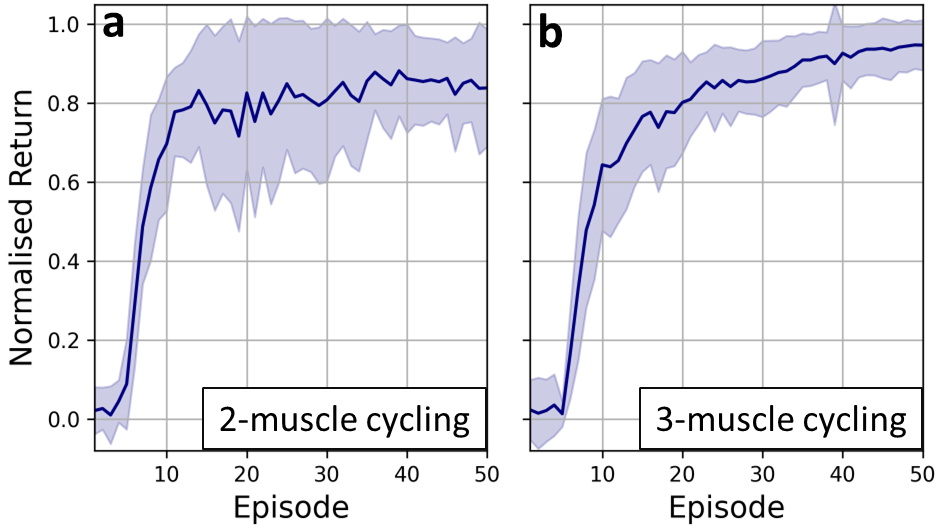}
    \vspace{-2mm}
    \caption{The model-based learning curves of (a) 2-muscle and (b) 3-muscle cycling. The solid curves and shades are the means and standard deviations over 10 configurations.
    \label{fig:returns}}
    \vspace{-4mm}
    \end{center}
\end{figure}

The model-based training progresses were monitored by episodic returns--the discounted reward that the RL agents obtained in each training episode. Fig.\ref{fig:returns}a and b show the normalised episodic returns of 2- and 3-muscle cases, respectively. In the early period of the training, the returns in both cases are in a near-zero region as the agents were unable to induce full-revolution cycling. After that, the returns rise quickly between the $5^{th}$ and $15^{th}$ episode as their abilities to induce the cycling motion improve. The rise is sharper in 2-muscle case as the 2-muscle patterns are easier to learn. The returns reach plateaus at around the $20^{th}$ episode, indicating the discoveries of optimal model-based patterns. Note that low return episodes, which result in large standard deviations, could occur even after the optimal patterns were discovered if the initial crank angles were close to the dead angles. The occurrences are more often in 2-muscle cases. The red bars in Fig.\ref{fig:pattern} is an example model-based pattern which was used in the real-world experiment to collect the data.
\vspace{-1mm}

\subsection{Real-world experiments}
We built an OpenSim cycling model corresponding to the real cycling setup (Fig.\ref{fig:Framework}c), trained an RL agent, and obtained the corresponding model-based pattern. The whole process took approximately 10 minutes; the measurement took roughly 2 minutes, and the RL training took around 8 minutes (on core i7-9700K). After that, ten 10-second real cycling sessions were carried out and the data were collected. This took approximately 5 minutes, including short breaks between sessions. The RL agent was trained in an offline manner on the collected data. This process took roughly 1 minute. Note that the model-based training took longer because it involved both computing the cycling simulation and updating RL policy (neural networks' parameters $\boldsymbol{\phi}$ and $\boldsymbol{\varphi}$), while the offline training involved only the policy update. On the EMG pattern side, we collected EMG from a 30-second voluntary cycling session at comfortable speed, around 60 RPM. We applied a threshold of 2V, an average level of movement noises, to determine the voluntary actives.

\begin{figure}[hb!]
    \begin{center}
\includegraphics[width=0.72\columnwidth]{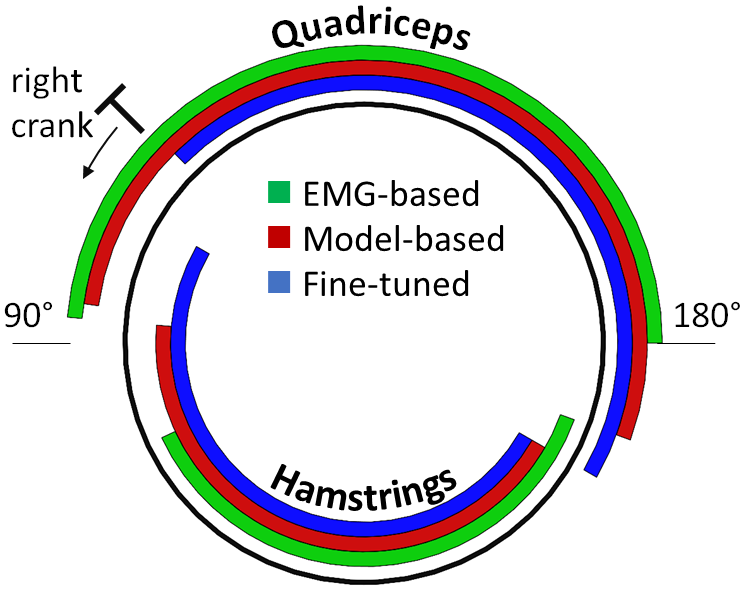}
    \vspace{-1mm}
    \caption{(a) The (green) EMG-based, (red) model-based, and (blue) fine-tuned stimulation patterns w.r.t the right crank moving counterclockwise.
    \label{fig:pattern}}
    \end{center}
    \vspace{-4mm}
\end{figure}

Fig.\ref{fig:pattern} compares the model-based, EMG-based, and fine-tuned patterns. The \emph{hamstrings} \emph{Off} angles are similar across different patterns. The fine-tuned pattern a has larger \emph{hamstrings} active range than the others. For \emph{quadriceps}, the model-based pattern has the largest active range. The \emph{quadriceps} \emph{Off} angles of EMG- and model-based patterns are very close, while the model-based and fine-tuned are similar in the \emph{On} angle. Noticeably, the EMG-based \emph{On} angles are behind those of the others. This reflects the muscles' activation delay for which the pattern has to compensate by starting the stimulation before the crank reaches the angles where the muscles produce positive torque.

\begin{figure}[hb!]
    \begin{center}
    \includegraphics[width=0.9\columnwidth]{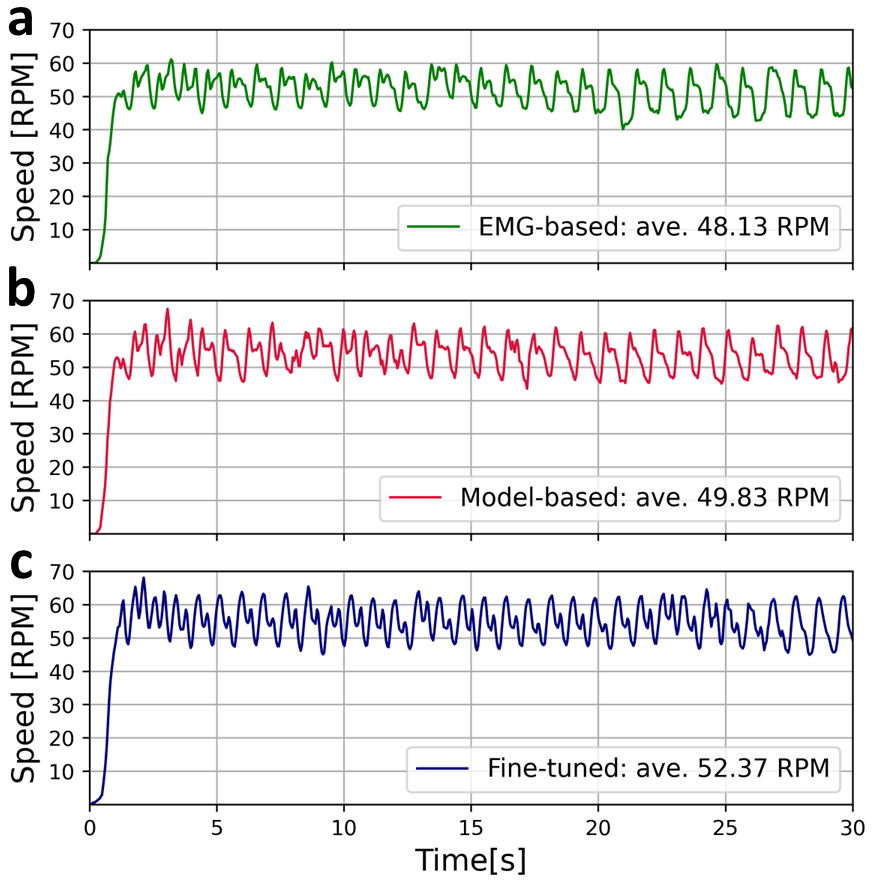}
    \vspace{-2mm}
    \caption{Cycling crank speed induced by (a) EMG-based, (b) model-based, and (c) fine-tuned patterns over 30-second sessions.
    \label{fig:speed}}
    \end{center}
    \vspace{-4mm}
\end{figure}

Fig.\ref{fig:speed}a-c compares the cycling performances of the three patterns in terms of the crank's speed. The out-of-the-box model-based pattern (Fig.\ref{fig:speed}b) successfully induced cycling motion on the real cycling system. The model-based pattern produced the average speed of 49.83 RPM which is slightly higher than that produced by the EMG-based pattern (Fig.\ref{fig:speed}a). The speed difference is mainly due to the differences in the \emph{On} angles, with the earlier \emph{On} results in faster speed. The fine-tuned pattern produced the highest average speed at 52.37 RPM (Fig.\ref{fig:speed}c).

\section{Discussion \& Conclusion}
We present an RL-based method for finding the stimulation pattern that requires neither force nor EMG sensors. Our method learns model-based patterns on customised neuromechanical models and uses cycling data to fine-tune the patterns through offline-RL. The evaluations show that our method can robustly find the patterns for different cycling configurations. The real-world experiment shows that the out-of-the-box model-based pattern can induce cycling motion with similar performance to that of the EMG-based pattern. The model-based pattern is improved by using the cycling data and offline RL, yielding the fine-tuned pattern that has the best performance in term of speed.


One limitation of the proposed method is that muscular fatigue can affect the cadence and therefore influence the fine-tuned pattern result. The fatigue effect can be partially addressed by applying weights on the cadence data collected from different trials to compensate for the fatigue. Alternatively, we have recently introduced a machine-learning-based method \cite{Wannawas2023b} that could be used to infer hidden states. Our method has yet to be tested on paralysed individuals. One challenge in such a case is that it requires assistance, either from motors or clinicians, to initiate cycling motions. The data collected during the assisted and inertia-influenced periods have to be properly processed or discarded. It is also worth to noting that, in this experiment, the roller resistance was set to be very low so that the results of good or bad stimulation can be observed immediately.


Regarding the algorithm itself, this method does not exploit the potential of using the data to optimise the OpenSim model which can lead to better model-based patterns and a shorter fine-tuning phase. It is also worth mentioning that the optimality itself is with respect to the objective or the reward function (Eq.\ref{eq:reward}). Our setup here is meant for achieving high-speed cycling. This may not always be the best, for example, for long cycling. Finding the pattern for other cycling purposes can be done by modifying the reward function such as increasing $\beta$ to obtain more stimulation-efficient patterns for long cycling. Regarding the performance in this experiment, the fine-tuned pattern's cycling speed is not much higher than that of the EMG-based pattern. This is partially because the stimulation current used in the experiment was quite low, making the differences less pronounced. It should be also highlighted that, outside laboratories, an EMG-based pattern is more difficult to obtain because it needs equipment and experimental setup. In contrast, our method only requires the data generated during cycling.

Crank angle measuring device also plays an important role in enabling the public use of our method. Although there are several ways to interface our method with the devices of existing cycling systems, building the interface can be a technical challenge for the users. Developing a low-cost, dedicated device that can be easily installed on the existing systems will tremendously benefit the users. In this regard, IMU-based devices strapped on the legs \cite{Wiesener2017,Schmoll2022,Maneski2017} are attractive because they can be built using inexpensive electronic components and do not require any modification on the existing cycling equipment.

In a broader view, we believe that artificial intelligence and electrical stimulation will have an important role in rehabilitation systems of the future. This proof-of-concept work is a case study of human-in-the-loop AI in rehabilitation, displaying its feasibility and potential. Specifically, this real-world success of reinforcement learning in FES control is a step towards the development on AI-based intelligent controls that can power rehabilitation robots for the restoration of general movements \cite{Shafti2019,Stewart2019,Wannawas2023a}.

\addtolength{\textheight}{-10cm}   


\section*{ACKNOWLEDGMENT}
NW acknowledges his support by the Royal Thai Government Scholarship. AAF acknowledges his support by UKRI Turing AI Fellowship (EP/V025449/1).

\bibliography{bib}
\bibliographystyle{IEEEtran}

\end{document}